\documentclass{article}

\usepackage[eandd, nonanonymous, preprint]{neurips_2026} 

\usepackage[utf8]{inputenc}
\usepackage[T1]{fontenc}   
\usepackage{hyperref}     
\usepackage{url}         
\usepackage{booktabs}      
\usepackage{amsfonts}   
\usepackage{nicefrac}    
\usepackage{microtype}
\usepackage{xcolor}        
\usepackage{graphicx}
\usepackage{geometry}
\usepackage{multirow}
\usepackage{tcolorbox}
\usepackage{xspace}
\usepackage{verbatim}
\usepackage{amsmath}
\usepackage{longtable, booktabs, array, enumitem}
\title{PulseLM: A Foundation Dataset and Benchmark for PPG-Text
Learning}

\author{%
Hung Manh Pham$^{1,}$\thanks{Equal contribution} , Jinyang Wu$^{1, *}$, Xiao Ma$^{1}$, Yiming Zhang$^{1}$, Yixin Xu$^{2}$, \\ \textbf{Aaqib Saeed$^{3}$}, \textbf{Bin Zhu$^{1,}$}\thanks{Corresponding authors} , \textbf{Zhou Pan$^{1,\dag}$}, \textbf{Dong Ma$^{4,\dag}$} \\
$^{1}$Singapore Management University \quad
$^{2}$Queen Mary University of London \quad \\
$^{3}$Eindhoven University of Technology \quad
$^{4}$University of Cambridge \\
\texttt{\{hm.pham.2023, jinyang.wu.2024, xiaoma.2022\}@smu.edu.sg} \\
\texttt{\{yimingzhang, binzhu, panzhou\}@smu.edu.sg} \\ \texttt{yixin.xu@qmul.ac.uk \quad a.saeed@tue.nl \quad dm878@cam.ac.uk} \\
}

\newcommand{\SysName}{PulseLM\xspace}

\begin{document}

\maketitle

\begin{abstract}
  Photoplethysmography (PPG) is a widely used non-invasive sensing modality for continuous cardiovascular and physiological monitoring across clinical, laboratory, and wearable settings. While existing PPG datasets support a broad range of downstream tasks, they typically provide supervision in the form of numerical measurements or task-specific labels, limiting their compatibility with language-based interfaces and multimodal foundation models. In this work, we introduce \SysName, a large-scale PPG-text question-answering dataset that bridges raw PPG waveforms and natural language through a unified question-answering (QA) formulation. \SysName aggregates PPG recordings from sixteen publicly available sources and harmonizes heterogeneous annotations into 12 downstream tasks. The dataset comprises over 1 million standardized 10-second PPG segments, associated with nearly 2.5 million question-answer pairs. We further define reproducible data pipeline, training, and evaluation protocols and establish baseline benchmarks using multimodal PPG-aware large language models. \SysName provides a standardized foundation for studying language-grounded physiological inference, cross-dataset generalization, and scalable benchmarking of PPG-based multimodal models. We publicly release the dataset and code at \href{https://huggingface.co/datasets/Manhph2211/PulseLM}{Hugging Face}
and \href{https://github.com/manhph2211/PULSE-LM}{GitHub}, respectively.

\end{abstract}

\section{Introduction}

Photoplethysmography (PPG) is a non-invasive optical sensing modality that underpins a wide range of modern health monitoring systems, from clinical pulse oximeters~\cite{VitalDB} to consumer wearables such as smartwatches and earbuds~\cite{applewatch, empatica, earbuds}. In principle, PPG measures changes in blood volume within the peripheral vasculature by emitting light onto the skin using a light-emitting diode (LED) and detecting variations in the intensity of transmitted or reflected light with a photodetector, where cardiac-induced pulsatile blood flow gives rise to a characteristic waveform. From that, PPG waveforms can be analyzed to enable continuous estimation of a broad range of cardiovascular and respiratory indicators, including heart rate (HR)~\cite{PPG-DALIA}, heart rate variability (HRV)~\cite{hu2025morphology,UTSA-PPG}, respiratory rate (RR)~\cite{Queensland-VS}, blood pressure (BP)~\cite{BCG}, stress levels~\cite{WESAD}, and sleep-disordered breathing (SDB)~\cite{SDB}. Owing to its low cost, ease of deployment, and applications, PPG has become one of the most widely collected physiological signals worldwide. 

From a machine learning perspective, this abundance of PPG data presents a significant opportunity for developing physiological models that can operate across tasks, devices, and recording environments. In fact, recent advances in deep pretrained PPG foundation models (e.g., Papagei and PulsePPG) have demonstrated their effectiveness on diverse PPG downstream tasks~\cite{pulseppg, papagei, abbaspourazad2024largescale}. In particular, these approaches leverage a pretrained single-modal PPG encoder to enable rapid and data-efficient development of downstream models. However, this paradigm still requires the post hoc development of task-specific models through linear probing or fine-tuning, while offering limited semantic interpretability and user interaction. 

In the context of modern multimodal foundation models, such as in domains like natural language processing and computer vision, large language models (LLMs) assisted models have demonstrated that language can serve as a unifying interface for reasoning, explanation, and cross-task generalization~\cite{sellergren2025medgemma, liu2023visual, plaat2025multi}. Specifically for physiological sensing, similar advances have been enabled by large-scale electrocardiogram (ECG)-text and image-report datasets, which allow models to align raw signals with clinical semantics expressed in natural language~\cite{liu2024teach, oh2023ecg, pham2025qheart, zhang2025sensorlm}, significantly improving the utility of these models. However, the development of such models requires signal-text-aligned datasets across different PPG downstream tasks, which are \textit{absent in the context of PPG research}.

One promising way to bridge this gap is to formulate PPG tasks as a single LLM-based question-answering (QA) problem. Question answering provides a natural and flexible interface that mirrors how clinicians and end users interact with physiological data, for example, by asking whether a heart rate is normal, whether signal quality is sufficient, or whether a recording indicates elevated cardiovascular risk. Importantly, a QA formulation can unify heterogeneous downstream tasks under a single supervision paradigm, avoiding the need for task-specific regression heads while enabling models to associate continuous waveform patterns with discrete, interpretable physiological concepts. 

However, enabling PPG-centric question answering at scale poses substantial challenges. Specifically, \textit{public PPG datasets are highly fragmented, differing in sensor placement} (fingertip vs. wrist vs. ear), \textit{sampling rate} (60 Hz vs. 128 Hz), \textit{recording environment} (clinical vs. in-the-wild, stationary vs. motion), \textit{subject population} (patients vs healthy, young vs. elderly people), and \textit{available annotations} (heart rate, stress level, motion status, etc.). \textit{Moreover, none of these datasets natively provides natural-language descriptions and QA supervision}. As a result, there is currently no large-scale, standardized benchmark that supports QA interfaces over PPG signals across diverse physiological domains and real-world conditions.

To address this issue, we introduce \SysName, the first large-scale PPG-Text QA dataset and benchmark designed to support multimodal PPG learning and interpretation. \SysName aggregates and harmonizes PPG recordings from diverse public sources spanning clinical, laboratory, and in-the-wild environments. For each standardized PPG segment, we derive physiologically grounded categorical labels from existing annotations or co-recorded reference signals (e.g., ECG) and convert them into QA supervision using controlled natural-language question-answer pools. The resulting dataset comprises over 1 million PPG-Text samples and nearly 2.5 million closed-ended question-answer pairs covering a broad range of rhythm, cardiovascular, stress, and respiratory tasks.

In addition to releasing the dataset, we establish a unified benchmarking protocol for PPG question answering and provide baseline results using multimodal models that integrate PPG encoders with instruction-tuned language models. By framing PPG interpretation as a language-based QA problem, \SysName enables systematic evaluation of unified physiological inference across tasks, datasets, and recording conditions, and lays the foundation for future research on multimodal foundation models for PPG signals.

This work makes the following contributions: 

\begin{itemize}
\vspace{-0.6pt}
    \item To the best of our knowledge, we introduce the first dataset to unify multiple heterogeneous PPG sources spanning clinical, laboratory, and in-the-wild environments into a single language-based QA benchmark. The dataset comprises over 1 million PPG samples and nearly 2.5 million question-answer pairs, covering 12 downstream tasks.
    \item We define corresponding benchmarks for multimodal PPG question-answering tasks, using consolidated PPG-language models that integrate pretrained PPG encoders with instruction-tuned LLMs to enable unified inference within a single QA framework.
    \item We release a unified data processing pipeline, model training, and evaluation protocols to facilitate future research on language-based PPG understanding.
\end{itemize}

\begin{figure*}[t]
  \centering
  \includegraphics[width=0.95\linewidth]{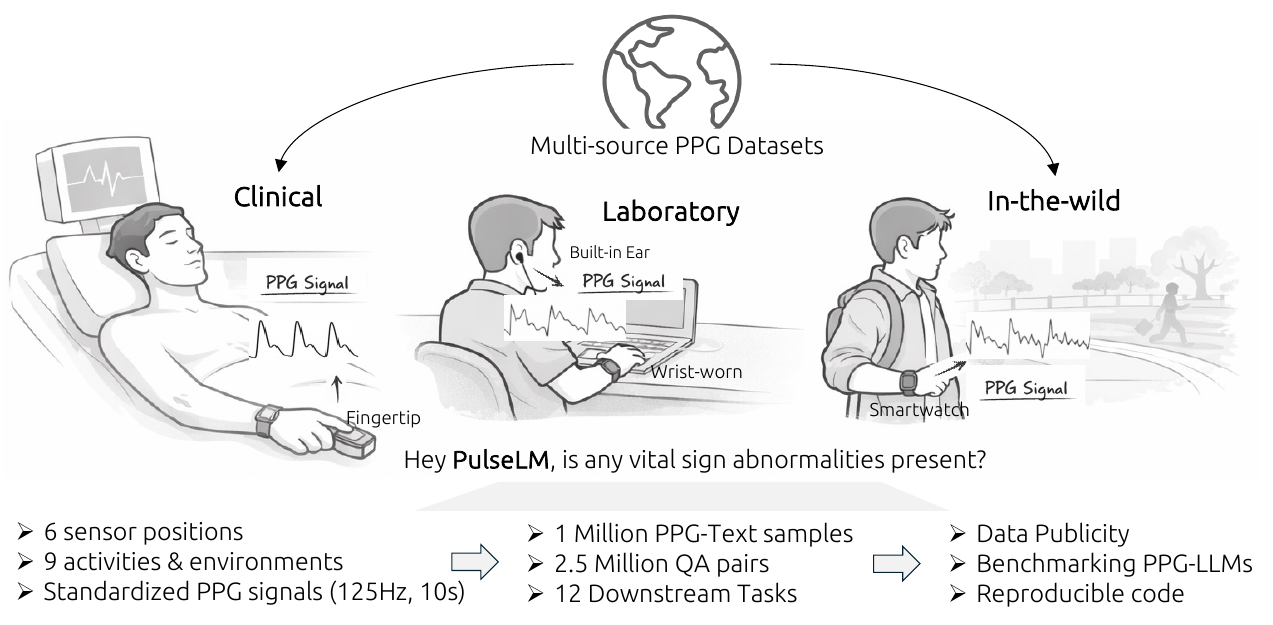}
\caption{Overview of our dataset study.}
\label{fig:overview}
\end{figure*} 

\section{Background}

\label{sec:background}

\subsection{PPG Downstream Tasks}
\label{sec:ppg_tasks}

PPG measures volumetric changes in blood within peripheral tissue by illuminating the skin with a light source and recording variations in the reflected or transmitted light. Owing to this optical measurement principle, the resulting waveform is inherently sensitive to both physiological conditions and sensing-related factors. Under stable clinical conditions, such as fingertip recordings with minimal motion, PPG typically exhibits quasi-periodic cardiac cycles with well-defined morphological components, including a dominant systolic peak and a secondary diastolic peak associated with vascular reflections. However, in practical wearable scenarios, factors such as user movement~\cite{Wang2019MAs} or a loose contact~\cite{ho2025wf} between a wrist-worn device and the skin can disrupt sensor–skin coupling and deform underlying tissue, leading to severe waveform distortion, attenuation of pulsatile components, or even the disappearance of characteristic peaks.

However, with the development of signal processing techniques~\cite{Peng2014AdaptiveFilter, Tang2016WaveletAndEMD}
and, more recently, deep learning methods~\cite{Zargari2023CycleGAN, kazemi2025respiration, PPG-DALIA, papagei, pham2025reliable}, PPG has enabled a broad spectrum of downstream tasks, including heart rate (HR), heart rate variability (HRV), and respiratory rate (RR) estimation, which are now routinely deployed in consumer wearables. These tasks rely primarily on temporal characteristics of the waveform and form the foundation of many real-world PPG-based systems. Beyond those core metrics, more complex inference tasks depend heavily on subtle waveform morphology. Blood pressure (BP) estimation~\cite{pan2024robust, gonzalez2023benchmark}, including both systolic and diastolic pressure, has been extensively studied using PPG-derived features/representation, either through regression to continuous values or classification into clinically meaningful categories (e.g., hypertension stage 1). 

In a related vein, rhythm-level cardiac inference, such as arrhythmia detection, exploits irregular inter-beat intervals, variable pulse amplitudes, and disrupted waveform morphology captured by PPG. These tasks require models to capture fine-grained morphological and rhythm structure and longer-range temporal dependencies within the signal. More recently, PPG has been explored in a range of non-traditional and higher-level inference domains. Studies have demonstrated its utility for emotion (stress) assessment~\cite{WESAD}, sleep-disordered breathing (SDB)~\cite{SDB} detection. Inference in these domains extends beyond purely signal-level pattern recognition and demands more expressive representations of physiological states.

\subsection{Existing PPG Datasets and Their Limitation}
\label{sec:ppg_dataset_limits}

To date, a wide range of publicly available PPG datasets have been released
over the past decades to support research in health monitoring and wearable sensing, spanning clinical, laboratory, and in-the-wild recording environments. For example, clinical datasets~\cite{VitalDB, UCI, BIDMC} are the most common category, collected in hospitals or clinics daily for a large number of patients, and typically use fingertip pulse oximeters to acquire high-fidelity PPG signals with minimal motion interference. Laboratory datasets extend this setting by adopting wearable form factors, most commonly wrist-worn or built-in ear devices~\cite{WESAD, EarSet}, and introducing structured protocols or controlled activities (e.g., mild physical movement such as sitting or talking) to study physiological responses under reproducible and specialized conditions. In contrast, in-the-wild or daily activity datasets~\cite{WildPPG, PPG-DALIA} further improve ecological validity by capturing longer-term PPG recordings during unconstrained everyday activities (e.g., running, playing sports) using consumer-grade wearables, at the cost of increased motion artifacts and sensing variability.

Despite the diversity of available PPG datasets and their applications, existing resources exhibit several fundamental limitations when viewed through the lens of unified physiological understanding. Most datasets are designed around a limited downstream task and provide supervision in the form of numerical values or categorical labels specific to that task, such as heart rate values (e.g., 65 bpm) for heart rate estimation tasks. Consequently, models trained on these datasets are typically optimized for narrow objectives and lack a unified interface for cross-task inference.

A second limitation arises from dataset fragmentation. Public PPG datasets differ substantially in sensor placement, sampling rate, signal length, preprocessing pipelines, and label definitions. Even for ostensibly similar tasks, such as heart rate variability estimation, different datasets rely on distinct metrics, time windows, or reference standards. This heterogeneity makes it difficult to combine datasets or evaluate models under consistent protocols, and often leads to models that implicitly overfit to dataset-specific characteristics rather than learning generalizable physiological representations.

\textit{Finally, and most critically for language-enabled modeling, existing open-source PPG datasets lack natural-language supervision}. Unlike ECG or medical imaging domains, where large-scale signal-text or image-report datasets~\cite{PhysioNet-mimic-iv-ecg-1.0, liu2024teach} have enabled models to learn clinical semantics through language, PPG datasets remain confined to numeric or concretized categorical labels. \textit{This absence of language-based supervision precludes the development of models that can describe waveform characteristics, explain physiological states, or interact with users through natural-language queries}. Taken together, these points suggest that progress in PPG-based learning is constrained not only by modeling approaches but also by the structure and scope of available datasets.

\section{~\SysName Dataset}
                          
\begin{table*}[t]
\centering
\setlength{\tabcolsep}{3pt}
\caption{Consolidated statistics of the \SysName dataset constructed from sixteen public PPG sources.}
\label{tab:qa_stats}
\resizebox{\textwidth}{!}{
\begin{tabular}{c l c c c l c l}
\toprule
\textbf{No.} &
\textbf{Dataset} &
\textbf{\# PPG Segments} &
\textbf{\# QA Cats} &
\textbf{\# QA Pairs} &
\textbf{QA Categories} &
\textbf{Sensor Positions} &
\textbf{Environment} \\
\midrule
(1)  & VitalDB~\cite{VitalDB}        & 163{,}959 & 5 & 819{,}441 & HR, BP, HRV                  & finger        & clinical \\
(2)  & UCI~\cite{kachuee2015cuff}            & 111{,}751 & 3 & 335{,}253 & HR, BP, SQI                  & finger        & clinical \\
(3)  & BCG~\cite{BCG}            & 671       & 3 & 2{,}013   & HR, BP, SQI                  & finger        & clinical \\
(4)  & PPG-BP~\cite{PPG-BP}         & 369       & 2 & 738       & HR, BP                       & finger        & clinical \\
(5)  & SDB~\cite{SDB}            & 258{,}897 & 1 & 258{,}897 & SDB                          & finger        & clinical \\
(6)  & Sensors~\cite{aguirre2021blood}        & 2{,}061   & 3 & 6{,}183   & HR, BP, SQI                  & finger        & clinical \\
(7)  & UQVitalSigns~\cite{Queensland-VS}   & 37{,}018  & 5 & 153,659 & HR, BP, SpO$_2$, RR, SQI     & finger        & clinical \\
(8) & PPGArrhythmia~\cite{PPGArrhythmia}  & 46{,}827  & 1 & 46{,}827  & Arrhythmia                   & finger        & clinical \\
(9) & MIMIC PERform~\cite{AFPPG}  & 4{,}196   & 1 & 4{,}196   & AF (Binary)                           & finger        & clinical \\
(10) & BIDMC~\cite{pimentel2016toward} & 12{,}462 & 2& 24{,}924 & SpO$_2$, RR     & finger        & clinical \\
(11)  & EarSet~\cite{EarSet}         & 1{,}776   & 1 & 1{,}776   & HR                           & ear           & lab \\
(12)  & UTSA-PPG~\cite{UTSA-PPG}       & 16{,}925  & 4 & 67{,}700  & HR, HRV                      & wrist, finger & lab \\
(13) & WESAD~\cite{WESAD}          & 2{,}998   & 1 & 2{,}998   & Stress                       & wrist         & lab \\

(14) & DALIA~\cite{PPG-DALIA}          & 39{,}216  & 1 & 39{,}216    & HR                           & wrist         & in-the-wild \\
(15) & WildPPG~\cite{WildPPG}       & 240{,}000 & 4 & 574{,}544 & HR, HRV                      & wrist, mix    & in-the-wild \\
(16) & AFPPGECG~\cite{AFPPGECG}       & 140{,}436 & 1 & 140{,}436  & AF (Binary)                             & wrist         & in-the-wild \\

\midrule
\textbf{Total} &
\textbf{16 source datasets} &
\textbf{1,079,562} &
\textbf{12} &
\textbf{2,478,801} &
 mix & mix & mix \\
\bottomrule
\end{tabular}}
\end{table*}

\begin{figure*}[t]

  \centering
  \includegraphics[width=0.97\linewidth]{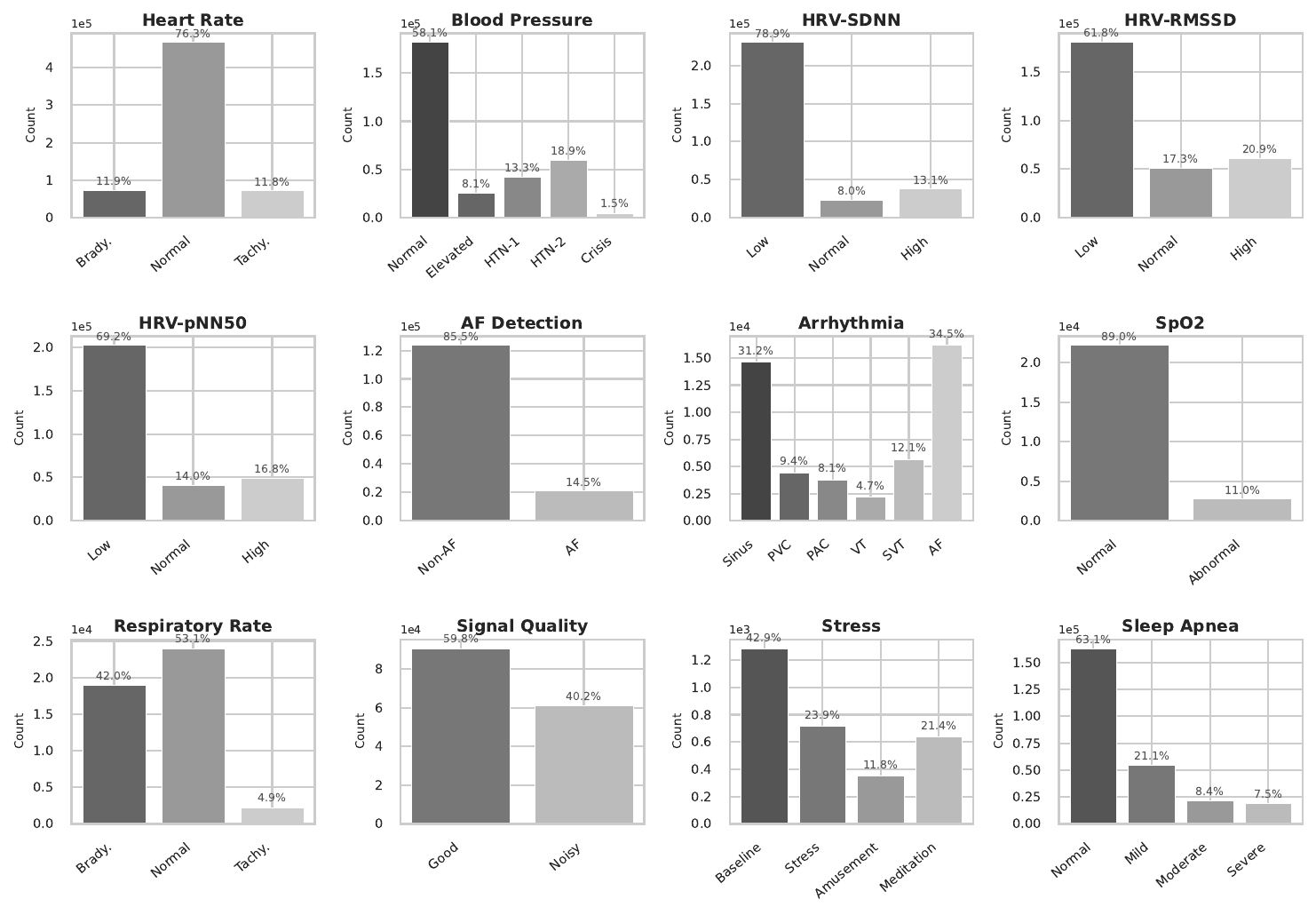}
\caption{Demonstration of label distributions in \SysName dataset.}
\label{fig:dist}
\end{figure*}

\label{sec:dataset}

\subsection{Overview}
\label{sec:dataset_overview}

We introduce \SysName, a large-scale PPG-Text question-answering dataset designed to enable language-based physiological understanding across diverse tasks and recording conditions (see Figure~\ref{fig:overview}). The dataset aggregates PPG recordings from 16 publicly available sources and reformulates their existing annotations into a unified QA representation. In total, \SysName contains 1,079,562 standardized PPG segments paired with 2,478,801 physiologically grounded question-answer pairs spanning twelve unique QA categories. Each PPG segment is represented as a 10-second window and may be associated with multiple QA pairs depending on the availability of annotations in the source dataset. By expressing heterogeneous downstream tasks under a shared QA interface, \SysName enables systematic benchmarking of multimodal models on unified physiological inference rather than task-specific prediction. To this end, the dataset spans clinical, laboratory, and in-the-wild wearable recordings, with PPG signals collected across diverse sensor placements such as finger pulse oximeters, wrist-worn wearables, in-ear sensors, and multi-site configurations. This diversity allows evaluation of robustness and generalization under realistic deployment scenarios. We will delve into key parts in the dataset below.

\subsection{Source Dataset Selection}
\label{sec:data_sources}
\SysName leverages PPG recordings from publicly available datasets selected to maximize overall dataset diversity. A consolidated summary of dataset-level statistics, including data size, sensor placement, and environment, is provided in Table~\ref{tab:qa_stats}. Across all sources, the dataset supports 12 physiological QA categories spanning heart rate, blood pressure, heart rate variability, atrial fibrillation (binary), arrhythmia, SpO$_2$, respiratory rate, sleep-disordered breathing, stress, and signal quality assessment. It further covers finger, wrist, ear, and other multi-location PPG acquisition, as well as a wide range of device characteristics. Taken together, this diversity is a deliberate design choice to discourage overfitting to dataset-specific artifacts and to promote learning of transferable physiological representations.

\subsection{Signal Standardization}
\label{sec:signal_standardization}

To enable consistent multimodal learning across heterogeneous data sources, all PPG recordings are standardized through a unified preprocessing pipeline comprising four stages: (1) \textit{Resampling}, where signals are resampled to a common rate of 125~Hz to eliminate variability across devices and protocols; (2) \textit{Filtering}, where a fourth-order Butterworth low-pass filter (8~Hz cutoff) is applied to suppress high-frequency noise, followed by DC (non-pulsatile) removal via per-segment mean subtraction to reduce baseline drift while preserving the hemodynamically relevant AC (pulsatile) component; (3) \textit{Segmentation}, where signals are partitioned into fixed-length 10-second windows, which are widely used to support PPG downstream tasks~\cite{papagei, pulseppg}; and (4) \textit{Normalization}, where each segment is independently scaled to the $[0,1]$ range using min-max normalization to mitigate amplitude variations while preserving waveform morphology and temporal structure.

\subsection{Ground Truth Harmonization}
\label{sec:ground_truth}

In our study, \SysName derives labels from original dataset annotations or from reference signals co-recorded with PPG, such as ECG. Specifically, depending on the source, available ground truth includes continuous-valued physiological measurements (e.g., blood pressure, SDB, and stress), categorical clinical labels (e.g., atrial fibrillation or arrhythmia type), contextual states (e.g., activity or stress), and signal quality indicators. For datasets that provide synchronized ECG recordings (e.g., VitalDB, WildPPG, and UTSA-PPG), HR and HRV labels (including ultra-short RMSSD: Root Mean Square of Successive Differences, SDNN: Standard Deviation of Normal-Normal intervals, and pNN50: Percentage of NN intervals differing by more than 50 ms) are computed using established ECG-based signal processing toolkits~\cite{Makowski2021neurokit}, treating ECG as the gold standard for rhythm and variability estimation. Ultimately, we provide the label distributions across tasks in Figure~\ref{fig:dist}.

Next, to enable unified language-based supervision across heterogeneous datasets, all ground truth is harmonized into a shared categorical label space using predefined mapping rules. Continuous-valued measurements are discretized into clinically or physiologically meaningful categories (e.g., bradycardia/normal/tachycardia for heart rate, standard risk-hypertension stages for blood pressure), including ECG-derived HR and HRV metrics where applicable. Dataset-provided categorical labels are preserved when compatible and otherwise normalized to a consistent label set. All mapping rules and thresholds are fixed \emph{a priori} and applied deterministically; the complete specifications are provided in Appendix~\ref{sec:label_mapping}. Through this process, \SysName supports twelve QA tasks spanning cardiovascular state, rhythm analysis, variability assessment, signal quality, behavior and affect, respiratory function, and sleep-related risk. All QA supervision is grounded in original annotations or reference signals, and language models infer no labels during dataset construction.

\subsection{Question Answering Formulation}
\label{sec:qa_formulation}

\SysName represents physiological supervision using a QA formulation, in which each PPG segment is paired with natural-language questions and answers. Depending on the availability of annotations in the source dataset, a single PPG segment may be associated with multiple QA instances spanning different physiological domains. For each QA category, a predefined answer vocabulary is specified based on the harmonized physiological labels described in Section~\ref{sec:ground_truth}. This formulation constrains model outputs to a finite and interpretable set of responses, enabling consistent supervision and objective evaluation across heterogeneous datasets.

To introduce linguistic variability while preserving semantic consistency, questions are generated for each QA category using controlled templates and paraphrasing. Specifically, we leverage OpenAI's GPT-5 with a strict instruction prompt to output candidate rephrasings, which are subsequently manually verified and all covered in Table~\ref{fig:questions}, to ensure semantic equivalence and quality. Furthermore, all answers are deterministically assigned by mapping existing ground-truth annotations to the corresponding categorical labels. As a result, the QA construction procedure does not introduce new or hallucinatory physiological information. Following that, to facilitate clear, reproducible benchmarking, we partition the data into train, validation, and test splits using subject-disjoint separation applied per source dataset, ensuring no subject overlap across splits while keeping the split ratio generally around 8/1/1. Importantly, paraphrases in the test split are strictly disjoint from those used during training. Full data examples and split statistics are provided in Appendix~\ref{appendix}.                   

\section{~\SysName Benchmarking}

\begin{figure*}[t]
  \centering
  \includegraphics[width=0.85\linewidth]{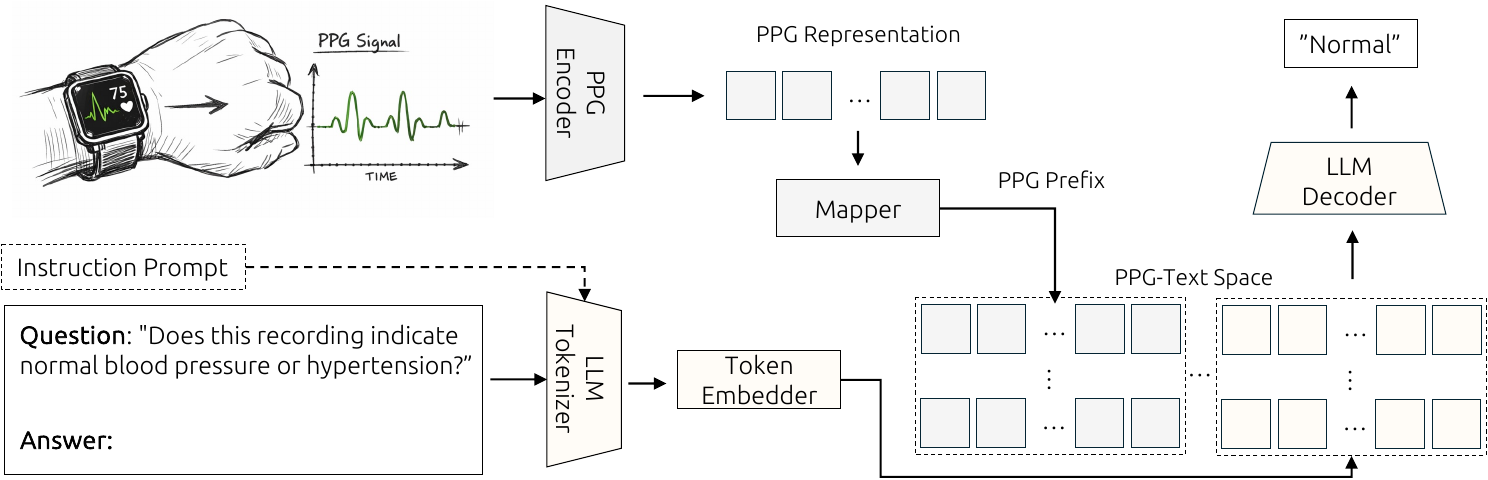}
\caption{Illustration of benchmarking PPG-LLM baselines.}
\label{fig:baseline}
\end{figure*}

\subsection{Objectives and Scope}
\label{sec:benchmark_overview}

Having established the \SysName dataset construction and QA formulation in Section~\ref{sec:dataset}, we now turn to the central empirical question: \emph{can multimodal models learn to ground natural-language physiological questions in raw PPG signals?} To this end, our goal is to provide an initial evaluation of language model behavior across diverse PPG dataset domains (e.g., sensor positions or recording environments) captured by \SysName.

Accordingly, our benchmarking pursues two complementary objectives. First, we establish reproducible in-domain baselines by training and evaluating models on matched subject-wise train-test splits from the same source datasets, providing a direct measure of how well current multimodal architectures can exploit the unified QA supervision that \SysName provides. Second, we employ cross-dataset generalization by training on a single source dataset and evaluating on entirely held-out sources with different sensor placements and recording environments. This setting directly tests whether models learn transferable physiological representations or merely overfit to dataset-specific signal characteristics.  Exact-match (EM) accuracy is used as the strict metric throughout for the text generation task, supplemented by per-dataset breakdowns to reveal where models succeed and where they fall short. All benchmarks operate over the harmonized label space described in Section~\ref{sec:dataset}, ensuring consistent evaluation across heterogeneous sources.

\subsection{Baseline Models}
\label{sec:benchmark_models}

To establish a comprehensive reference for performance on \SysName, we benchmark two complementary families of models: non-LLM transformer baselines that isolate the contribution of individual modalities, and full multimodal PPG-language models that form the primary focus of this benchmark.

\paragraph{Non-LLM Baselines.}
Before attributing performance to multimodal grounding, it is essential to quantify what each modality contributes independently as natural baselines for our tasks. Inspired by the experimental design in ECG-QA~\cite{oh2023ecg}, we include non-LLM baselines that systematically vary modality access. First, the Blind model processes only the natural-language question through a frozen BERT encoder~\cite{bert} and a linear classification head, without any access to the PPG signal. The Deaf model takes the complementary perspective: it operates exclusively on the PPG signal via a pretrained PPG encoder and a linear classifier, receiving no textual input. Lastly, the Fusion model integrates both modalities without an LLM: a PPG encoder and a BERT text encoder produce independent representations, which are concatenated and passed through a multi-layer perceptron to predict the answer. All baselines predict over a shared global answer vocabulary across all 12 QA categories. Together, they constitute an interpretable modality-ablation suite, enabling us to determine whether LLM-based models exploit physiological signals or rely on language shortcuts.

\paragraph{Multimodal PPG-Language Models.}
In addition to those baselines, we adopt a unified architecture, illustrated in Figure~\ref{fig:baseline}, that couples a pretrained PPG encoder with an instruction-tuned LLM via a lightweight projection layer, enabling joint modeling of physiological signals and textual queries without modifying the LLM structure. Specifically, a PPG encoder first processes the 10-second input waveform, producing a 512-dim latent embedding that encapsulates physiological features. A linear projection layer then maps this embedding into the LLM's token embedding space, yielding a PPG prefix token that is prepended to the embedded question tokens. The resulting multimodal sequence is fed into the LLM decoder, which attends jointly over both modalities to generate the answer.

A central design choice in this architecture is the PPG encoder, as it determines the quality of physiological features passed to the language model. This choice is particularly important given the heterogeneity of PPG signals across datasets (e.g., sensor placement, motion artifacts, and acquisition conditions), which imposes strong demands on the encoder’s ability to learn robust, transferable representations. Therefore, we employ two well-established pretrained PPG encoders: Papagei~\cite{papagei}, trained on large-scale PPG data at 125\,Hz, and PulsePPG~\cite{pulseppg}, pretrained at 50\,Hz (signals are resampled accordingly). Pairing each LLM backbone with both common encoders allows us to examine the effect of physiological representation quality from that of language model capacity.

For the LLM backbone, we evaluate a set of models: Qwen3-4B, Qwen2.5-7B-Instruct, LLaMA3.2-3B-Instruct, and LLaMA3.1-8B-Instruct. This selection balances model capacity with computational feasibility and provides an initial investigation into how model scale and backbone design affect physiological grounding performance using the proposal \SysName dataset.

\subsection{Implementation Details}
\label{sec:benchmark_implementation}

To ensure fair and reproducible comparison across all model variants, we adopt a unified training protocol for all models. Because the pretrained PPG encoders already capture rich physiological representations, we freeze their parameters throughout fine-tuning, adapting only the projection layer and the language model. For language model adaptation, we employ low-rank adaptation (LoRA)~\cite{lora} with rank $r=8$ and scaling factor $\alpha=16$ (dropout $0.1$). The projection layer is trained with a slightly higher learning rate of $2\times10^{-4}$ to encourage faster alignment, while LoRA parameters use $1\times10^{-4}$. Regarding the instruction prompt, we present it in detail from Box~\ref{box:pulselm_prompt}. All models are optimized with AdamW using a cosine learning rate schedule with a linear warm-up over the first 3\% of training steps, and trained for 2 epochs with a batch size of 16 and 4-step gradient accumulation in bfloat16 mixed precision on a single NVIDIA H200 GPU (140GB VRAM). For evaluation, all models use greedy decoding with a maximum of 32 new tokens. Predicted answers are extracted by parsing the <answer> tag in the generated output; if no valid tag is found or the text does not match any option, the prediction is treated as incorrect. All hyperparameters, training scripts, and evaluation code are released alongside the dataset.

\subsection{Main Results}
\label{sec:benchmark_results}

\begin{table*}[t]
  \centering
  \small    
  \setlength{\tabcolsep}{4pt}                                                                      
  \caption{In-domain evaluation using EM accuracy across different datasets. Columns correspond to individual datasets (1)-(16)
  following the order in Table~\ref{tab:qa_stats}.}                                                                                                         \label{tab:benchmark_dataset}                             
  \resizebox{\textwidth}{!}{                                                                                                                                     
  \begin{tabular}{l l c c c c c c c c c c c c c c c c c}    
  \toprule
  \textbf{Language Models} &                                                                                                                                               
  \textbf{PPG Encoder} &
  \textbf{(1)} & \textbf{(2)} & \textbf{(3)} & \textbf{(4)} & \textbf{(5)} &                                                                                     
  \textbf{(6)} & \textbf{(7)} & \textbf{(8)} & \textbf{(9)} & \textbf{(10)} &                                                                                    
  \textbf{(11)} & \textbf{(12)} & \textbf{(13)} & \textbf{(14)} & \textbf{(15)} & \textbf{(16)} &
  \textbf{Avg. $\pm$ Std} \\                                                                                                                                               
  \midrule 

% \midrule

  Blind PPG        & None     & 52.6 & 54.8 & 58.3 & 62.8 & 24.1 & 66.0 & 61.0 & 28.9 & 48.8 & 55.4 & 88.5 & 52.1 & 18.3 & 83.7 & 65.9 & 86.1 & 56.7 $\pm$ 19.8 \\
  \midrule
  \multirow{2}{*}{Deaf PPG}
                     & Papagei  & 49.0 & 42.4 & 17.2 & 42.3 & 25.8 & 45.3 & 52.8 & 13.0 & 50.8 & 87.9 & 79.9 & 44.5 & 0.0  & 61.8 & 35.0 & 81.2 & 45.5 $\pm$ 23.8 \\
                     & PulsePPG & 46.8 & 44.8 & 40.1 & 66.7 & 51.0 & 44.5 & 47.8 & 42.6 & 9.6  & 61.3 & 45.6 & 25.8 & 1.7  & 51.0 & 48.3 & 53.3 & 42.6 $\pm$ 16.5 \\
  \midrule
  \multirow{2}{*}{Fusion PPG}
                     & Papagei  & 71.9 & 60.5 & 73.4 & 60.3 & 36.4 & 67.6 & 64.3 & 28.5 & 45.2 & 84.2 & 72.3 & 53.1 & 23.3 & 65.9 & 70.0 & 86.1 & 60.2 $\pm$ 17.9 \\
                     & PulsePPG & 55.8 & 71.2 & 74.0 & 47.4 & 43.0 & 73.3 & 71.3 & 65.6 & 44.4 & 69.9 & 75.3 & 59.1 & 41.5 & 68.5 & 76.6 & 85.7 & 63.9 $\pm$ 13.2 \\
  \midrule

  \multirow{2}{*}{Qwen3-4B}      & Papagei  & 77.1 & 52.9 & 74.0 & 69.2 & 61.0 & 48.0 & 66.7 & 48.5 & 49.9 & 88.2 & 88.5 & 63.9 & 39.5 & 83.7 & 85.3 & 86.1 & 67.7 $\pm$ 15.9 \\
                                & PulsePPG & 74.8 & 68.5 & 32.3 & 69.2 & 54.9 & 72.4 & 75.9 & 52.9 & 65.8 & 87.9 & 89.6 & 65.7 & 44.4 & 84.5 & 85.4 & 85.2 & 69.3 $\pm$ 16.0 \\     \midrule

  \multirow{2}{*}{LLaMA3.2-3B-Instruct}   & Papagei  & 77.6 & 47.6 & 74.0 & 69.2 & 61.0 & 42.9 & 75.0 & 48.4 & 49.9 & 88.2 & 88.5 & 63.9 & 39.5 & 83.7 & 85.3 & 86.1 & 67.6 $\pm$ 16.8 \\
& PulsePPG & 75.0 & 69.0 & 38.5 & 69.2 & 55.9 & 71.2 & 74.9 & 51.4 & 71.8 & 86.6 & 90.1 & 65.9 & 43.7 & 85.1 & 85.2 & 84.9 & 69.9 $\pm$ 15.1 \\ \midrule

  \multirow{2}{*}{LLaMA3.1-8B-Instruct}    & Papagei  & 77.9 & 53.6 & 74.0 & 69.2 & 61.0 & 54.3 & 75.0 & 42.7 & 49.9 & 88.2 & 88.5 & 63.9 & 39.5 & 83.7 & 85.3 & 86.1 & 68.3 $\pm$ 16.0 \\
& PulsePPG & 74.7 & 69.4 & 49.5 & 69.2 & 58.4 & 70.8 & 74.3 & 49.6 & 75.0 & 81.3 & 89.6 & 65.7 & 47.4 & 85.0 & 85.2 & 84.5 & 70.6 $\pm$ 13.1 \\  \midrule

  \multirow{2}{*}{Qwen2.5-7B-Instruct}      & Papagei  & 77.5 & 55.5 & 64.6 & 69.2 & 61.0 & 49.5 & 70.9 & 48.4 & 49.9 & 68.5 & 88.5 & 63.9 & 39.5 & 83.7 & 85.3 & 86.1 & 66.4 $\pm$ 14.7 \\
& PulsePPG & 75.4 & 77.8 & 79.2 & 69.2 & 54.1 & 75.1 & 72.9 & 59.6 & 62.9 & 75.6 & 90.7 & 65.1 & 49.7 & 86.2 & 85.1 & 85.9 & 72.8 $\pm$ 11.6 \\

  \bottomrule
  \end{tabular}
  }                                                         
  \end{table*}

\textbf{In-domain evaluation.}
Table~\ref{tab:benchmark_dataset} reports results across all 16 datasets. First, a clear performance gap can be observed between non-LLM baselines and multimodal PPG-language models. The non-LLM approaches, including Blind, Deaf, and Fusion PPG models, consistently achieve lower performance, with average accuracy generally below 63\%. This suggests that relying on either language or signal alone, or combining them through simple feature fusion, is insufficient for capturing the full range of physiological semantics in \SysName. In contrast, multimodal PPG-language models deliver generally stronger results across datasets. On average, these models improve performance by approximately 6\%-9\%, reaching around 70\% accuracy. In particular, Qwen2.5-7B-Instruct paired with PulsePPG achieves over 72\% average accuracy, representing a substantial gain over all non-LLM baselines.

We also observe that across multimodal LLM architectures with different PPG encoders, PulsePPG generally yields better performance than Papagei, and both outperform non-LLM baselines. This suggests two points. First, integrating pretrained PPG encoders with language models is an effective design choice, as both encoders lead to meaningful performance gains within this framework. Second, the advantage of PulsePPG may stem from its pretraining on more diverse datasets spanning both laboratory and in-the-wild settings, including various levels of motion-affected samples, whereas Papagei is primarily trained on more controlled clinical data, which may limit its robustness under heterogeneous conditions. Another important factor in our framework is the choice of LLM backbone. Although all LLM-based models outperform non-LLM baselines, performance varies across architectures. Models based on the Qwen family tend to achieve slightly better overall results compared to LLaMA, indicating that backbone design and pretraining strategy also influence multimodal physiological inference performance.

\begin{table*}[t]
  \centering
  \footnotesize
  \caption{Cross-dataset evaluation: models trained on source datasets and evaluated on held-out datasets. The PPG encoder used is PulsePPG (except blind), and results are reported with 95\% bootstrap confidence intervals.}
  \label{tab:cross_dataset}
  \resizebox{\textwidth}{!}{
  \begin{tabular}{l ccc ccc}
  \toprule
  & \multicolumn{3}{c}{\textbf{VitalDB}}
  & \multicolumn{3}{c}{\textbf{WildPPG}} \\
  \cmidrule(lr){2-4} \cmidrule(lr){5-7}
  \textbf{Model}
  & BCG & UTSA-PPG & EarSet
  & DALIA & UTSA-PPG & EarSet \\
  \midrule
  Blind PPG  & $60.9_{\scriptstyle\pm8.5}$ & $44.4_{\scriptstyle\pm0.9}$ & $88.5_{\scriptstyle\pm3.3}$ & $65.9_{\scriptstyle\pm1.3}$ & $59.1_{\scriptstyle\pm0.9}$ & $72.3_{\scriptstyle\pm4.6}$ \\   \midrule
  Deaf PPG   & $43.0_{\scriptstyle\pm8.6}$ & $34.0_{\scriptstyle\pm0.9}$ & $47.0_{\scriptstyle\pm5.1}$ & $68.6_{\scriptstyle\pm1.2}$ & $46.6_{\scriptstyle\pm0.9}$ & $79.7_{\scriptstyle\pm4.1}$ \\   \midrule
  Fusion PPG & $60.9_{\scriptstyle\pm8.5}$ & $44.4_{\scriptstyle\pm0.9}$ & $86.0_{\scriptstyle\pm3.6}$ & $68.7_{\scriptstyle\pm1.2}$ & $59.4_{\scriptstyle\pm0.9}$ & $73.4_{\scriptstyle\pm4.5}$ \\
  \midrule
  Qwen2.5-7B-Instruct  & $60.9_{\scriptstyle\pm8.5}$ & $49.5_{\scriptstyle\pm0.9}$ & $89.0_{\scriptstyle\pm3.2}$ & $85.3_{\scriptstyle\pm1.0}$ & $63.9_{\scriptstyle\pm0.9}$ & $88.5_{\scriptstyle\pm3.3}$ \\
  \bottomrule
  \end{tabular}
  }
  % \vspace{-0.2pt}
\end{table*}

\textbf{Cross-dataset evaluation.}
In this section, we present additional analysis of cross-dataset generalization, focusing on model transfer across datasets with varying populations, devices, and acquisition conditions. Table~\ref{tab:cross_dataset} evaluates two representative cross-domain scenarios using VitalDB and WildPPG as training datasets, and compares our best-performing model (i.e., Qwen2.5-7B-Instruct with PulsePPG) against predefined non-LLM baselines. In the VitalDB scenario, models are trained on the clinical finger-sensor data and evaluated on wearable datasets with distinct sensor placements or environments (EarSet, BCG, UTSA-PPG). This setting investigates robustness to cross-dataset variation when transferring from controlled clinical data to more diverse conditions. Similarly, in the WildPPG scenario, models are trained on large-scale in-the-wild wrist recordings and evaluated on other recording conditions (DALIA, UTSA-PPG, EarSet). This setting examines transfer from noisy, naturalistic data to more controlled environments.

Across both scenarios, we observe trends generally consistent with the in-domain evaluation, with the multimodal LLM-based approach consistently outperforming the non-LLM baselines. Despite substantial differences in sensor placement and acquisition conditions, the models maintain relatively competitive performance across several held-out datasets, suggesting that the learned representations capture transferable physiological information beyond dataset-specific and participant-specific patterns. At the same time, the variability across transfer settings indicates that cross-domain generalization in PPG remains sensitive to dataset heterogeneity. Overall, these findings reinforce the role of \SysName as a benchmark for systematically studying representation transfer and robustness across diverse physiological sensing environments.

\section{Conclusion}

This paper presents \SysName, a large-scale dataset designed to support multimodal physiological inference from PPG signals using language-based supervision. Consolidating heterogeneous PPG recordings from multiple sources into a unified closed-ended question-answering task, \SysName enables consistent modeling and evaluation across diverse downstream inferences. The accompanying baseline results provide an initial reference, highlighting both the promise and current challenges of multimodal physiological LLMs under realistic data heterogeneity, task diversity, and domain shifts. We hope \SysName to serve as a reference point for future work on scalable physiological foundation models over wearable biosignals.

\bibliographystyle{ACM-Reference-Format}
\bibliography{references}

\clearpage
% \onecolumn
\appendix

\section{Additional Dataset Details} \label{appendix}

This appendix provides supplementary material to support further dataset and benchmark details for reproducibility and transparency. Specifically, we include illustrative PPG waveform samples (Appendix~\ref{sec:waveform_examples}), the label harmonization rules (Appendix~\ref{sec:label_mapping}), the full question templates per category (Appendix~\ref{sec:question_templates}), the dataset structure (Appendix~\ref{sec:dataset_structure}), and dataset split statistics (Appendix~\ref{sec:split_stats}). We present each of them in the sections below.

\subsection{PPG Waveform Examples}
\label{sec:waveform_examples}

\begin{figure*}[t]

  \centering
  \includegraphics[width=\linewidth]{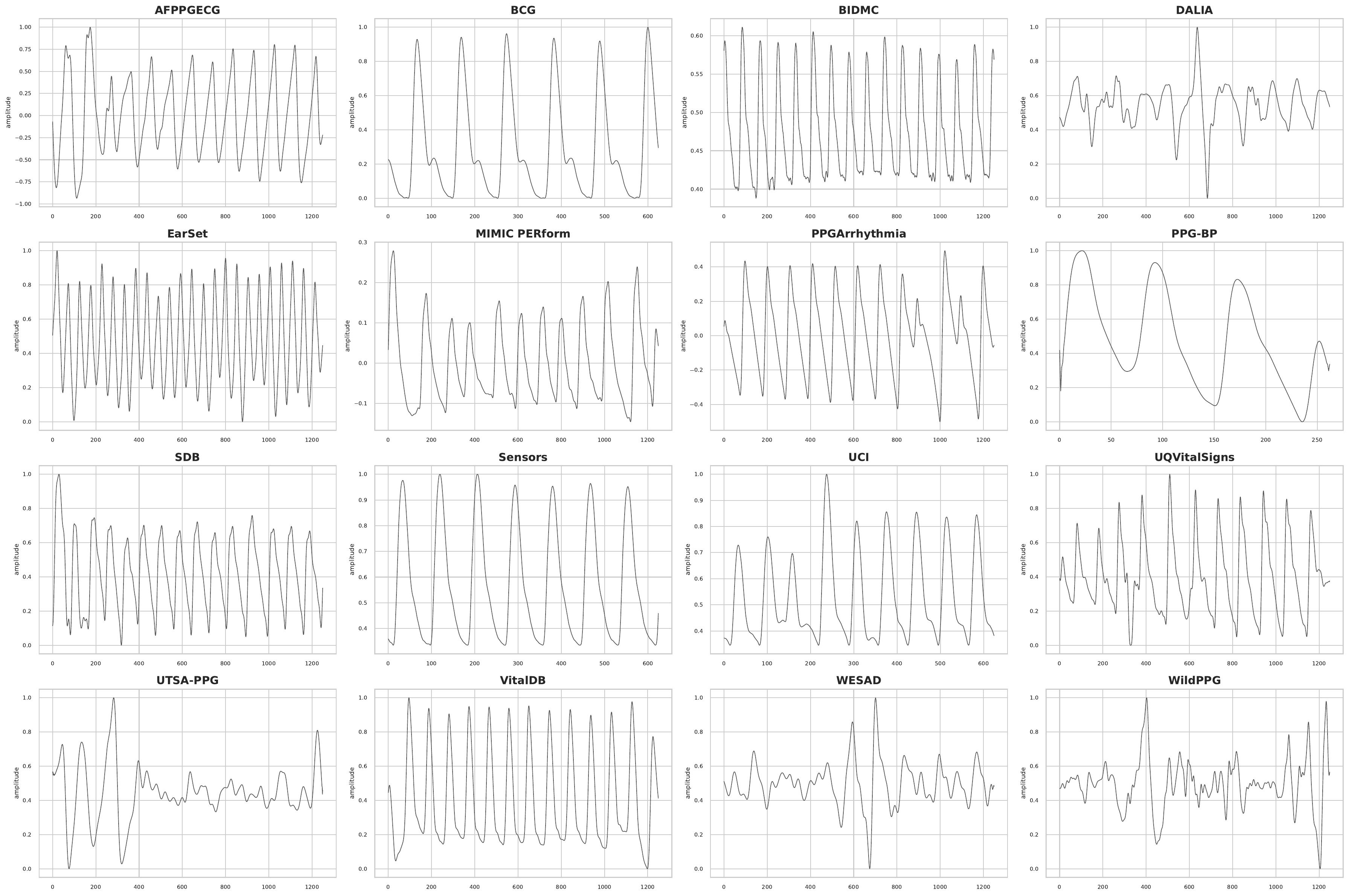}
\caption{Illustrative PPG waveforms from the \SysName train dataset. Each recording is windowed to 10\,s; shorter signals are padded to match this fixed duration.}                               
\label{fig:examples}
\end{figure*}

\begin{figure*}[t]

  \centering
  \includegraphics[width=\linewidth]{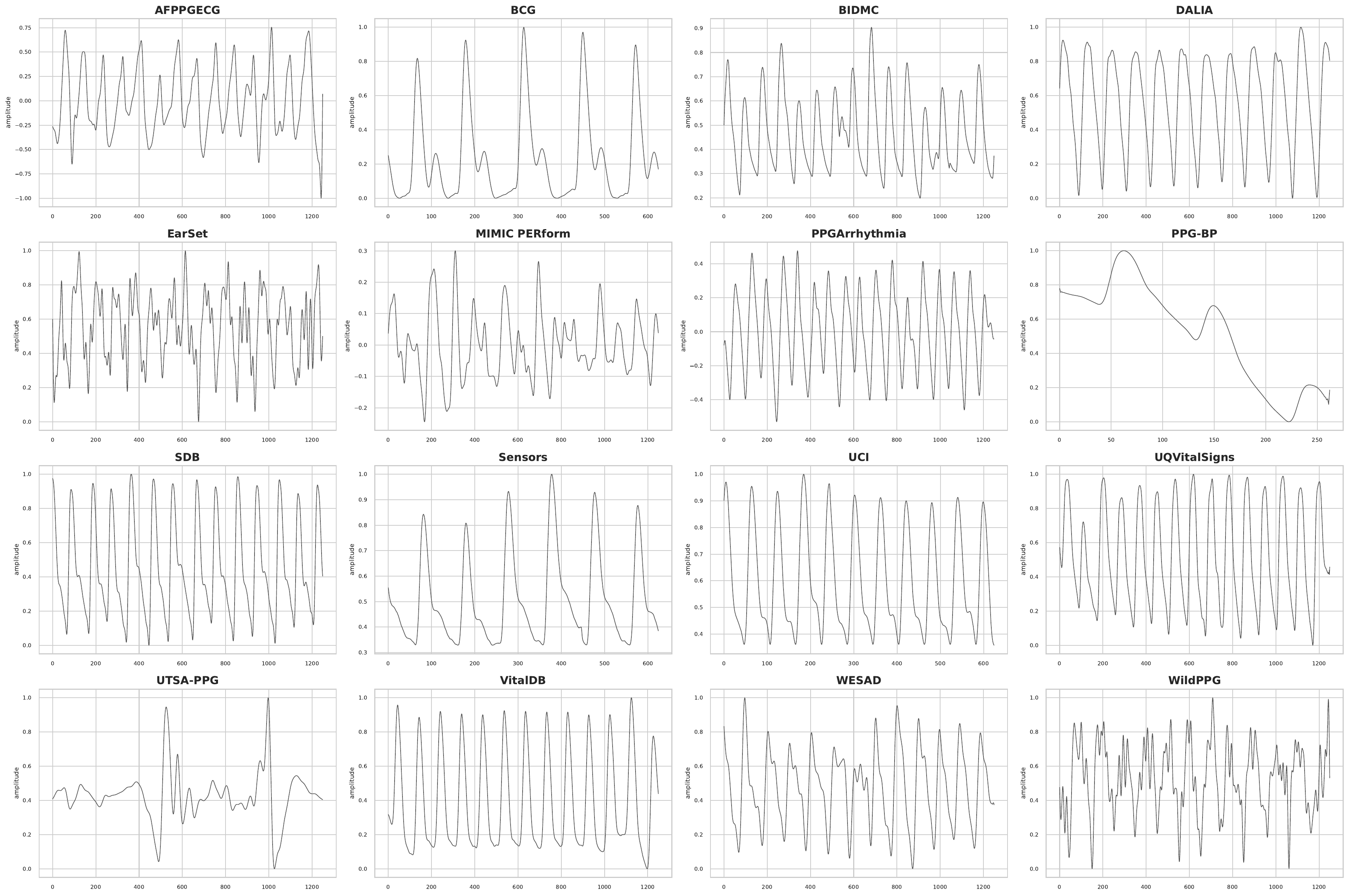}
\caption{Illustrative PPG waveforms from the \SysName test dataset. Each recording is windowed to 10\,s; shorter signals are padded to match this fixed duration.}                               
\label{fig:examples2}
\end{figure*}

First, Figures~\ref{fig:examples} and~\ref{fig:examples2} show representative 10-second PPG waveforms from the train and test partitions of \SysName. The examples span multiple source datasets and sensor placements, illustrating the considerable morphological diversity in the benchmark. Clinical finger-sensor recordings (e.g., VitalDB, BCG) tend to exhibit well-defined systolic peaks, while wrist-worn and in-ear recordings (e.g., WildPPG, Dalia) show broader, more attenuated morphologies. Signals captured during unconstrained activity exhibit visible motion artifacts and baseline drift, highlighting the challenge of cross-environment generalization.

\subsection{Label Mapping}
\label{sec:label_mapping}

In our study, as the included PPG sources provide heterogeneous annotations (e.g., continuous numerical values such as blood pressure), we harmonize all physiological measurements into a unified set of categorical labels. The mapping rules are summarized in Table~\ref{tab:label_mapping}, which was derived from established clinical guidelines (e.g, AHA, AASM) where available and complemented with commonly used practices and literature to enable consistent large-scale benchmarking. 

\begin{table*}[t]
\centering
\renewcommand{\arraystretch}{1.25}
\setlength{\tabcolsep}{10pt}
\small
\caption{Label mapping rules for physiological signal harmonization.}
\label{tab:label_mapping}
\begin{tabular}{p{3.2cm} p{9cm}}
\toprule
\textbf{Category} & \textbf{Mapping rule} \\
\midrule

\textbf{Heart rate} &
HR (bpm) is discretized into three classes: \texttt{bradycardia} ($<60$), \texttt{normal} ($60$--$100$), and \texttt{tachycardia} ($>100$). \\

\textbf{Blood pressure} &
SBP/DBP (mmHg) are mapped using severity-prioritized rules: \texttt{hypertensive crisis} (SBP$>180$ or DBP$>120$), \texttt{stage 2} (SBP$\ge140$ or DBP$\ge90$), \texttt{stage 1} (SBP$\ge130$ or DBP$\ge80$), \texttt{elevated} (SBP 120--129 and DBP$<80$), and \texttt{normal} (SBP$<120$ and DBP$<80$). \\

\textbf{Signal quality} &
Skewness-based rule: $0.5 \le \text{skew} \le 2.0$ is \texttt{good\_quality}; otherwise \texttt{noisy}. \\

\textbf{Sleep apnea} &
AHI (events/hour) is mapped as \texttt{normal} ($<5$), \texttt{mild} ($5$--$14$), \texttt{moderate} ($15$--$29$), and \texttt{severe} ($\ge30$). \\

\textbf{HRV} &
RMSSD (ms): \texttt{low} ($<20$), \texttt{normal} ($20$--$50$), \texttt{high} ($>50$). SDNN (ms): \texttt{low} ($<50$), \texttt{normal} ($50$--$100$), \texttt{high} ($>100$). pNN50 (\%): \texttt{low} ($<3$), \texttt{normal} ($3$--$25$), \texttt{high} ($>25$). \\

\textbf{SpO$_2$} &
SpO$_2$ (\%) is categorized as \texttt{normal} ($\ge95$), \texttt{abnormal}. \\

\textbf{Respiratory rate} &
RR (breaths/min) is classified as \texttt{bradypnea} ($<12$), \texttt{normal} ($12$--$20$), and \texttt{tachypnea} ($>20$). \\

\bottomrule
\end{tabular}
\end{table*}

\subsection{Question Templates}
\label{sec:question_templates}

Importantly, we report the full set of question paraphrases for each QA category in Table~\ref{fig:questions}, alongside the prompts used for generation (Box~\ref{box:paraphrase_prompt}) with GPT-5. Each category uses 10 templates for training/validation and 5 strictly disjoint paraphrases for testing, ensuring evaluation measures linguistic generalization rather than template memorization.

 \begin{tcolorbox}[
      colframe=black,
      title=\footnotesize{LLM Prompt for Question Template Generation per QA Category.},
      % auto counter,
      label={box:paraphrase_prompt}
  ]                                                         
  \small
                                                            
  You are helping to build a PPG question-answering benchmark dataset. Based on the QA category and option below, generate 15 distinct paraphrased question templates.\\ \textbf{Input.}
  \textit{Category:} \texttt{\$\$} 
  \textit{Options:} \texttt{\$\$} \\
  \textbf{Output.} \texttt{} \\
  \end{tcolorbox}

\begin{longtable}{p{0.13\textwidth} p{0.42\textwidth} p{0.38\textwidth}}
\caption{Questions per category used during training/validation and test.} \\
  \toprule
\label{fig:questions}

\textbf{Category} & \textbf{Train \& Validation Questions} & \textbf{Test Questions} \\
\midrule
\endfirsthead
\multicolumn{3}{c}{\tablename\ \thetable{} -- continued from previous page} \\
\toprule
\textbf{Category} & \textbf{Train \& Validation Questions} & \textbf{Test Questions} \\
\midrule
\endhead
\midrule
\multicolumn{3}{r}{\textit{Continued on next page}} \\
\endfoot
\bottomrule
\endlastfoot

AF Detection&
\begin{minipage}[t]{6.2cm}\begin{enumerate}[leftmargin=*,topsep=0pt,itemsep=0pt,parsep=0pt]\footnotesize
  \item Assess whether atrial fibrillation is present in this PPG.
  \item Classify this PPG as AF or non-AF.
  \item Determine whether this signal indicates atrial fibrillation.
  \item Does this PPG signal show atrial fibrillation?
  \item Does this waveform indicate AF?
  \item Is atrial fibrillation present in this recording?
  \item Is this a normal rhythm or atrial fibrillation?
  \item Provide the atrial fibrillation detection result.
  \item What is the AF detection result for this segment?
  \item What is the AF label for this PPG recording?
\end{enumerate}\end{minipage} &
\begin{minipage}[t]{5.5cm}\begin{enumerate}[leftmargin=*,topsep=0pt,itemsep=0pt,parsep=0pt]\footnotesize
  \item Based on this PPG, report the rhythm status regarding AF.
  \item Give the atrial fibrillation screening outcome for this window.
  \item How would you characterize the cardiac rhythm in this recording?
  \item Select the rhythm label: AF or sinus rhythm.
  \item State the AF classification result for this signal.
\end{enumerate}\end{minipage} \\
\midrule

Arrhythmia &
\begin{minipage}[t]{6.2cm}\begin{enumerate}[leftmargin=*,topsep=0pt,itemsep=0pt,parsep=0pt]\footnotesize
  \item Assess the arrhythmia classification from this recording.
  \item Categorize the heart rhythm abnormality.
  \item Classify the cardiac rhythm in this recording.
  \item Determine the rhythm classification for this waveform.
  \item Identify the arrhythmia type from the signal.
  \item Is this a normal rhythm or arrhythmia?
  \item What cardiac rhythm category does this sample belong to?
  \item What is the arrhythmia category for this segment?
  \item What is the rhythm diagnosis for this segment?
  \item What type of arrhythmia does this signal show?
\end{enumerate}\end{minipage} &
\begin{minipage}[t]{5.5cm}\begin{enumerate}[leftmargin=*,topsep=0pt,itemsep=0pt,parsep=0pt]\footnotesize
  \item Give the arrhythmia assessment for this PPG window.
  \item How would you characterize the rhythm disorder in this recording?
  \item Name the cardiac rhythm pattern shown in this signal.
  \item Select the most fitting rhythm label for this segment.
  \item State the rhythm diagnosis based on this waveform.
\end{enumerate}\end{minipage} \\
\midrule

Blood Pressure &
\begin{minipage}[t]{6.2cm}\begin{enumerate}[leftmargin=*,topsep=0pt,itemsep=0pt,parsep=0pt]\footnotesize
  \item Based on the PPG, what is the BP category?
  \item Categorize the blood pressure level.
  \item Classify the blood pressure level shown in this PPG segment.
  \item Determine the BP classification from this waveform.
  \item Does this sample indicate normal blood pressure or hypertension?
  \item Provide the blood pressure risk category.
  \item What blood pressure class does this sample belong to?
  \item What hypertension stage does this PPG correspond to?
  \item What is the blood pressure category for this sample?
  \item What is the blood pressure status for this recording?
\end{enumerate}\end{minipage} &
\begin{minipage}[t]{5.5cm}\begin{enumerate}[leftmargin=*,topsep=0pt,itemsep=0pt,parsep=0pt]\footnotesize
  \item Give the blood pressure assessment for this recording.
  \item How would you characterize the blood pressure in this signal?
  \item Name the hypertension stage indicated by this PPG.
  \item Select the BP classification that applies to this sample.
  \item State the blood pressure diagnosis for this waveform.
\end{enumerate}\end{minipage} \\
\midrule

Heart Rate &
\begin{minipage}[t]{6.2cm}\begin{enumerate}[leftmargin=*,topsep=0pt,itemsep=0pt,parsep=0pt]\footnotesize
  \item Based on the PPG waveform, what is the HR category?
  \item Categorize the heart rate shown in this recording.
  \item Classify the heart rate based on this waveform.
  \item Determine the heart rate category from the signal.
  \item Is the heart rate normal, bradycardic, or tachycardic?
  \item Provide the clinical heart rate category.
  \item What heart rate classification does this PPG indicate?
  \item What is the heart rate category for this PPG segment?
  \item What is the heart rate status for this sample?
  \item Which heart rate class does this sample belong to?
\end{enumerate}\end{minipage} &
\begin{minipage}[t]{5.5cm}\begin{enumerate}[leftmargin=*,topsep=0pt,itemsep=0pt,parsep=0pt]\footnotesize
  \item Give the heart rate assessment for this recording.
  \item How would you characterize the heart rate in this signal?
  \item Name the heart rate condition shown in this PPG.
  \item Select the HR classification that applies to this sample.
  \item State the heart rate diagnosis for this waveform.
\end{enumerate}\end{minipage} \\
\midrule

HRV-pNN50 &
\begin{minipage}[t]{6.2cm}\begin{enumerate}[leftmargin=*,topsep=0pt,itemsep=0pt,parsep=0pt]\footnotesize
  \item Assess the pNN50 category from this PPG.
  \item Categorize the pNN50 variability measure.
  \item Classify the pNN50 level.
  \item Determine the pNN50 level for this recording.
  \item How would you categorize pNN50 for this PPG?
  \item Is pNN50 low, normal, or high in this sample?
  \item Provide the pNN50 category.
  \item What is the pNN50 category for this segment?
  \item What is the pNN50-based HRV classification?
  \item What pNN50 class does this sample belong to?
\end{enumerate}\end{minipage} &
\begin{minipage}[t]{5.5cm}\begin{enumerate}[leftmargin=*,topsep=0pt,itemsep=0pt,parsep=0pt]\footnotesize
  \item Give the pNN50 assessment for this recording.
  \item How would you characterize the pNN50 in this signal?
  \item Name the pNN50 condition shown in this PPG.
  \item Select the pNN50 classification that applies to this sample.
  \item State the pNN50 variability level for this waveform.
\end{enumerate}\end{minipage} \\
\midrule

HRV-RMSSD &
\begin{minipage}[t]{6.2cm}\begin{enumerate}[leftmargin=*,topsep=0pt,itemsep=0pt,parsep=0pt]\footnotesize
  \item Assess the RMSSD-based variability level.
  \item Categorize the short-term HRV (RMSSD).
  \item Classify the RMSSD-based heart rate variability.
  \item Determine the RMSSD classification.
  \item How would you categorize RMSSD here?
  \item Is the RMSSD low, normal, or high?
  \item Provide the RMSSD category for this sample.
  \item What RMSSD class does this recording indicate?
  \item What is the HRV RMSSD category for this segment?
  \item What is the parasympathetic activity level (RMSSD)?
\end{enumerate}\end{minipage} &
\begin{minipage}[t]{5.5cm}\begin{enumerate}[leftmargin=*,topsep=0pt,itemsep=0pt,parsep=0pt]\footnotesize
  \item Give the RMSSD assessment for this recording.
  \item How would you characterize the RMSSD in this signal?
  \item Name the RMSSD condition shown in this PPG.
  \item Select the RMSSD classification that applies to this sample.
  \item State the short-term HRV (RMSSD) level for this waveform.
\end{enumerate}\end{minipage} \\
\midrule

HRV-SDNN &
\begin{minipage}[t]{6.2cm}\begin{enumerate}[leftmargin=*,topsep=0pt,itemsep=0pt,parsep=0pt]\footnotesize
  \item Assess the SDNN-based variability category.
  \item Categorize the overall HRV (SDNN) level.
  \item Classify the SDNN-based heart rate variability level.
  \item Determine the SDNN level for this recording.
  \item How would you categorize SDNN for this PPG?
  \item Is the SDNN low, normal, or high in this sample?
  \item Provide the SDNN category based on this PPG segment.
  \item What SDNN class does this sample belong to?
  \item What is the HRV SDNN category for this segment?
  \item What is the SDNN-based HRV classification?
\end{enumerate}\end{minipage} &
\begin{minipage}[t]{5.5cm}\begin{enumerate}[leftmargin=*,topsep=0pt,itemsep=0pt,parsep=0pt]\footnotesize
  \item Give the SDNN assessment for this recording.
  \item How would you characterize the SDNN in this signal?
  \item Name the SDNN condition shown in this PPG.
  \item Select the SDNN classification that applies to this sample.
  \item State the overall HRV (SDNN) level for this waveform.
\end{enumerate}\end{minipage} \\
\midrule

Respiratory Rate &
\begin{minipage}[t]{6.2cm}\begin{enumerate}[leftmargin=*,topsep=0pt,itemsep=0pt,parsep=0pt]\footnotesize
  \item Assess the respiratory rate category from this PPG.
  \item Categorize the breathing rate.
  \item Classify the respiratory rate level.
  \item Determine the respiratory rate category for this recording.
  \item How would you categorize the respiratory rate here?
  \item Is the respiratory rate normal, slow, or fast?
  \item Provide the respiratory rate category.
  \item What is the breathing rate classification for this segment?
  \item What is the respiratory rate category for this sample?
  \item What respiratory rate class does this sample belong to?
\end{enumerate}\end{minipage} &
\begin{minipage}[t]{5.5cm}\begin{enumerate}[leftmargin=*,topsep=0pt,itemsep=0pt,parsep=0pt]\footnotesize
  \item Give the respiratory rate assessment for this recording.
  \item How would you characterize the breathing rate in this signal?
  \item Name the respiratory condition shown in this PPG.
  \item Select the RR classification that applies to this sample.
  \item State the respiratory rate diagnosis for this waveform.
\end{enumerate}\end{minipage} \\
\midrule

Sleep Apnea &
\begin{minipage}[t]{6.2cm}\begin{enumerate}[leftmargin=*,topsep=0pt,itemsep=0pt,parsep=0pt]\footnotesize
  \item Assess the respiratory disturbance category.
  \item Categorize the breathing disorder level.
  \item Classify the sleep breathing pattern.
  \item Determine the sleep apnea severity.
  \item Does this segment indicate sleep apnea?
  \item Provide the SDB classification for this PPG window.
  \item What is the AHI-based severity category?
  \item What is the breathing disorder category?
  \item What is the sleep-disordered breathing label for this segment?
  \item What sleep-disordered breathing class is this?
\end{enumerate}\end{minipage} &
\begin{minipage}[t]{5.5cm}\begin{enumerate}[leftmargin=*,topsep=0pt,itemsep=0pt,parsep=0pt]\footnotesize
  \item Give the sleep-disordered breathing assessment for this recording.
  \item How would you characterize the breathing disorder in this signal?
  \item Name the sleep apnea severity shown in this PPG.
  \item Select the SDB classification that applies to this sample.
  \item State the AHI-based severity for this waveform.
\end{enumerate}\end{minipage} \\
\midrule

SpO2 &
\begin{minipage}[t]{6.2cm}\begin{enumerate}[leftmargin=*,topsep=0pt,itemsep=0pt,parsep=0pt]\footnotesize
  \item Assess the oxygen saturation level from this PPG.
  \item Categorize the SpO2 level.
  \item Classify the blood oxygen saturation.
  \item Determine the SpO2 category for this recording.
  \item How would you categorize SpO2 for this PPG?
  \item Is the SpO2 normal or does it indicate hypoxemia?
  \item Provide the oxygen saturation category.
  \item What SpO2 class does this sample belong to?
  \item What is the SpO2 category for this segment?
  \item What is the oxygen saturation classification?
\end{enumerate}\end{minipage} &
\begin{minipage}[t]{5.5cm}\begin{enumerate}[leftmargin=*,topsep=0pt,itemsep=0pt,parsep=0pt]\footnotesize
  \item Give the SpO2 assessment for this recording.
  \item How would you characterize the oxygen saturation in this signal?
  \item Name the hypoxemia condition shown in this PPG.
  \item Select the SpO2 classification that applies to this sample.
  \item State the blood oxygen saturation diagnosis for this waveform.
\end{enumerate}\end{minipage} \\
\midrule

Signal Quality &
\begin{minipage}[t]{6.2cm}\begin{enumerate}[leftmargin=*,topsep=0pt,itemsep=0pt,parsep=0pt]\footnotesize
  \item Assess the signal quality of this PPG recording.
  \item Classify the PPG signal quality based on skewness.
  \item Determine the signal quality index category.
  \item How would you categorize the signal quality here?
  \item Is this PPG recording of good or poor quality?
  \item Is this PPG signal clean or motion distorted?
  \item Provide the SQI quality category for this sample.
  \item Rate the quality of this PPG signal.
  \item What is the SQI classification for this segment?
  \item What is the signal quality category for this PPG waveform?
\end{enumerate}\end{minipage} &
\begin{minipage}[t]{5.5cm}\begin{enumerate}[leftmargin=*,topsep=0pt,itemsep=0pt,parsep=0pt]\footnotesize
  \item Give the signal quality assessment for this recording.
  \item How would you characterize the PPG signal integrity here?
  \item Name the signal quality condition shown in this waveform.
  \item Select the SQI classification that applies to this sample.
  \item State the signal quality diagnosis for this PPG.
\end{enumerate}\end{minipage} \\
\midrule

Stress &
\begin{minipage}[t]{6.2cm}\begin{enumerate}[leftmargin=*,topsep=0pt,itemsep=0pt,parsep=0pt]\footnotesize
  \item Categorize the stress condition for this sample.
  \item Classify the stress level from this PPG.
  \item Determine the emotional/stress state.
  \item Identify the stress level for this segment.
  \item Provide the stress state for this PPG window.
  \item What is the affective state for this recording?
  \item What is the emotional state label?
  \item What is the stress label for this segment?
  \item What psychological state does this segment indicate?
  \item What stress category does this sample belong to?
\end{enumerate}\end{minipage} &
\begin{minipage}[t]{5.5cm}\begin{enumerate}[leftmargin=*,topsep=0pt,itemsep=0pt,parsep=0pt]\footnotesize
  \item Give the stress assessment for this recording.
  \item How would you characterize the psychological state in this signal?
  \item Name the stress condition shown in this PPG.
  \item Select the stress classification that applies to this sample.
  \item State the affective state diagnosis for this waveform.
\end{enumerate}\end{minipage} \\ \\

\label{tab:question_paraphrases}
\end{longtable}

\subsection{Dataset Structure and QA Format}
\label{sec:dataset_structure}

Each sample in \SysName is stored as a row in a Parquet file and contains three key fields: 1) \texttt{signal}: a fixed-length array of 1{,}250 floats representing a 10-second PPG segment sampled at 125\,Hz and normalized to $[0,1]$; 2) \texttt{qa}: a JSON object mapping each applicable question category to a \{\texttt{question}, \texttt{answer}\} pair; and 3) \texttt{text}, an optional natural-language description of the recording context, which is derived from original dataset metadata, such as physiological labels, demographic attributes, activity context, sensor placement, and recording environment (Table~\ref{tab:text_examples}). The \texttt{qa} field may contain multiple entries (i.e., 1-5) per sample, depending on which task labels are available for the recording. Box~\ref{box:qa_example} illustrates the schema of a \SysName sample in our Hugging Face repository, showing how a single PPG segment is associated with multiple QA pairs across distinct physiological categories.

\begin{tcolorbox}[
    colframe=black,
    title=\footnotesize{Sample schema in \SysName.},
    label={box:qa_example}
]
\small
\textbf{signal:} \texttt{[}$x_1, x_2, \ldots, x_{1250}$\texttt{]} \\[3pt]
\textbf{text:} \textit{"$\langle$...$\rangle$"} \\[3pt]
\textbf{qa:} \\[2pt]
\hspace*{4mm}\texttt{\{} \\
\hspace*{8mm}\texttt{"$\langle$category$_1\rangle$": \{ "question": "$\langle$...$\rangle$", "answer": "$\langle$...$\rangle$" \},} \\
\hspace*{8mm}\texttt{"$\langle$category$_2\rangle$": \{ "question": "$\langle$...$\rangle$", "answer": "$\langle$...$\rangle$" \},} \\
\hspace*{8mm}\texttt{"$\langle$category$_3\rangle$": \{ "question": "$\langle$...$\rangle$", "answer": "$\langle$...$\rangle$" \},} \\
\hspace*{8mm}\texttt{$\vdots$} \\
\hspace*{4mm}\texttt{\}}\\[3pt]
% \textbf{\textit{$\langle$metadata$\rangle$}} \textit{"$\langle$...$\rangle$"} 
\end{tcolorbox}

\subsection{Dataset Split Statistics}
\label{sec:split_stats}

To provide a more comprehensive view of our dataset, Tables~\ref{tab:ppg-counts}-\ref{tab:env-counts} report detailed split statistics across four dimensions: PPG segment counts per source dataset (Table~\ref{tab:ppg-counts}), QA pair counts by question category (Table~\ref{tab:qa-cat-counts}), QA pair counts per source dataset (Table~\ref{tab:qa-ds-counts}), and segment counts by recording environment (Table~\ref{tab:env-counts}). VitalDB and WildPPG are the two largest contributors. We can also observe that clinical recordings represent 59\% of the total data, and heart rate has the highest QA count while stress is the scarcest, reflecting the natural prevalence of these physiological states in the source datasets.

\begin{table}[h]
  \centering
  \caption{Number of PPG signals per dataset and split.}
  \label{tab:ppg-counts}
  \setlength{\tabcolsep}{15pt}
  \begin{tabular}{lrrrr}
    \toprule
    \textbf{Dataset} & \textbf{Train} & \textbf{Val} & \textbf{Test} & \textbf{Total} \\
    \midrule
      VitalDB & 131,324 & 16,485 & 16,150 & 163,959 \\
      UCI & 89,054 & 11,286 & 11,411 & 111,751 \\
      BCG & 521 & 86 & 64 & 671 \\
      PPG-BP & 294 & 36 & 39 & 369 \\
      SDB & 205,618 & 23,525 & 29,754 & 258,897 \\
      Sensors & 1,631 & 180 & 250 & 2,061 \\
      UQVitalSigns & 25,202 & 2,703 & 9,113 & 37,018 \\
      PPGArrhythmia & 36,820 & 4,764 & 5,243 & 46,827 \\
      MIMIC PERform & 3,239 & 240 & 717 & 4,196 \\
      BIDMC & 9,412 & 1,652 & 1,398 & 12,462 \\
      EarSet & 1,368 & 44 & 364 & 1,776 \\
      UTSA-PPG & 12,740 & 1,437 & 2,748 & 16,925 \\
      WESAD & 2,104 & 297 & 597 & 2,998 \\
      DALIA & 29,294 & 4,602 & 5,320 & 39,216 \\
      WildPPG & 180,000 & 15,000 & 45,000 & 240,000 \\
      AFPPGECG & 64,725 & 43,868 & 31,843 & 140,436 \\
    \midrule
      \textbf{Total} & 793,346 & 126,205 & 160,011 & 1,079,562 \\
    \bottomrule
  \end{tabular}
\end{table}

\begin{table}[h]
  \centering
  \caption{Number of QA samples per category and split.}
  \setlength{\tabcolsep}{15pt}
  \label{tab:qa-cat-counts}
  \begin{tabular}{lrrrr}
    \toprule
    \textbf{Category} & \textbf{Train} & \textbf{Val} & \textbf{Test} & \textbf{Total} \\
    \midrule
      AF Detection & 67,964 & 44,108 & 32,560 & 144,632 \\
      Arrhythmia & 36,820 & 4,764 & 5,243 & 46,827 \\
      Blood Pressure & 247,816 & 30,742 & 34,549 & 313,107 \\
      HRV-RMSSD & 226,828 & 23,677 & 41,945 & 292,450 \\
      HRV-SDNN & 226,828 & 23,677 & 41,945 & 292,450 \\
      HRV-pNN50 & 226,828 & 23,677 & 41,945 & 292,450 \\
      Heart Rate & 470,920 & 51,859 & 90,459 & 613,238 \\
      Respiratory Rate & 31,570 & 4,181 & 9,556 & 45,307 \\
      Signal Quality & 116,408 & 14,255 & 20,838 & 151,501 \\
      Sleep Apnea & 205,618 & 23,525 & 29,754 & 258,897 \\
      SpO2 & 17,889 & 2,563 & 4,492 & 24,944 \\
      Stress & 2,104 & 297 & 597 & 2,998 \\
    \midrule
      \textbf{Total} & 1,877,593 & 247,325 & 353,883 & 2,478,801 \\
    \bottomrule
  \end{tabular}
\end{table}

\begin{table}[h]
  \centering
  \caption{Number of QA samples per dataset and split.}
  \setlength{\tabcolsep}{15pt}
  \label{tab:qa-ds-counts}
  \begin{tabular}{lrrrr}
    \toprule
    \textbf{Dataset} & \textbf{Train} & \textbf{Val} & \textbf{Test} & \textbf{Total} \\
    \midrule
      VitalDB & 656,344 & 82,386 & 80,711 & 819,441 \\
      UCI & 267,162 & 33,858 & 34,233 & 335,253 \\
      BCG & 1,563 & 258 & 192 & 2,013 \\
      PPG-BP & 588 & 72 & 78 & 738 \\
      SDB & 205,618 & 23,525 & 29,754 & 258,897 \\
      Sensors & 4,893 & 540 & 750 & 6,183 \\
      UQVitalSigns & 106,031 & 11,515 & 36,113 & 153,659 \\
      PPGArrhythmia & 36,820 & 4,764 & 5,243 & 46,827 \\
      MIMIC PERform & 3,239 & 240 & 717 & 4,196 \\
      BIDMC & 18,824 & 3,304 & 2,796 & 24,924 \\
      EarSet & 1,368 & 44 & 364 & 1,776 \\
      UTSA-PPG & 50,960 & 5,748 & 10,992 & 67,700 \\
      WESAD & 2,104 & 297 & 597 & 2,998 \\
      DALIA & 29,294 & 4,602 & 5,320 & 39,216 \\
      WildPPG & 428,060 & 32,304 & 114,180 & 574,544 \\
      AFPPGECG & 64,725 & 43,868 & 31,843 & 140,436 \\
    \midrule
      \textbf{Total} & 1,877,593 & 247,325 & 353,883 & 2,478,801 \\
    \bottomrule
  \end{tabular}
\end{table}

\begin{table}[h]
  \centering
  \caption{Number of PPG segments by recording environment and split.}
  \label{tab:env-counts}
  \setlength{\tabcolsep}{15pt}
  \begin{tabular}{lrrrr}
    \toprule
    \textbf{Environment} & \textbf{Train} & \textbf{Val} & \textbf{Test} & \textbf{Total} \\
    \midrule
      Clinical & 503,115 & 60,957 & 74,139 & 638,211 \\
      Lab & 16,212 & 1,778 & 3,709 & 21,699 \\
      In-the-wild & 274,019 & 63,470 & 82,163 & 419,652 \\
    \midrule
      \textbf{Total} & 793,346 & 126,205 & 160,011 & 1,079,562 \\
    \bottomrule
  \end{tabular}
\end{table}

\begin{table}[t]
\centering
\caption{Representative text field examples from \SysName.}
\label{tab:text_examples}
\begin{tabular}{cp{0.82\linewidth}}
\toprule
  \multicolumn{1}{c}{\textbf{Example}} & \multicolumn{1}{c}{\textbf{Text Description}} \\                                                                                                     \midrule
 1 & The patient is a 55-year-old female with a height of 164 cm and a weight of 56 kg, resulting in a body mass index (BMI) of 20.8, which is within the normal range. She has olive skin and engages in physical exercise for approximately 5 hours per week. Currently, her activity level is classified as unknown. The patient's heart rate is recorded at 67 beats per minute, which falls within the normal heart rate category. \\
\midrule
 2 & The participant is a 27-year-old male with brown (Type V) skin tone. An in-ear PPG signal was recorded from the right ear using a green LED during a running activity. The heart rate was categorized as tachycardia, indicating a heart rate greater than 100 bpm. The recording occurred under full-body movement conditions, which may have introduced expected motion artifacts. \\
\midrule
 3 & The patient presents with tachycardia, with a heart rate exceeding 100 bpm. Blood pressure readings indicate stage 1 hypertension, characterized by a systolic blood pressure within the range of 130-139 mmHg and a diastolic blood pressure between 80-89 mmHg. The quality of the signal is noted as noisy or distorted, which may impact the reliability of the readings. Further evaluation and monitoring may be necessary to assess the patient's cardiovascular status. \\
\midrule
 4 & The patient exhibits bradycardia with a heart rate recorded at 57 bpm. Blood pressure measurements indicate a normal range. Signal quality is assessed as good. Further evaluation of the cardiac cycle reveals distinct phases, although specific duration measurements are not included in the verified data. Continuous monitoring is recommended to observe any changes in heart rate and overall cardiac function. \\
\midrule
 5 & The patient is a 77-year-old female with a height of 150 cm and a weight of 42 kg, resulting in a BMI of 18.7. Her blood pressure is recorded at 182/80 mmHg, categorizing her in a hypertensive crisis due to the systolic blood pressure exceeding 180 mmHg. Additionally, her heart rate is noted at 101 bpm, indicating tachycardia as it is greater than 100 bpm. There is no significant medical history reported. \\
\midrule
 6 & A 26-year-old male, with a height of 181 cm and weight of 75 kg, resulting in a BMI of 22.9, is currently engaged in meditation. The patient's emotional state is noted as being associated with meditation, indicating a potential focus on stress reduction or relaxation techniques. \\
\bottomrule
\end{tabular}
\end{table}

\section{Additional Training Details}

Table~\ref{tab:benchmark_main} presents in-domain results across different QA categories. Overall, multimodal LLM-based models consistently outperform the non-LLM baselines across most tasks, demonstrating the effectiveness of integrating pretrained PPG representations with language models for unified physiological inference. In particular, strong performance is observed for tasks such as HR or AF detection, where the underlying physiological patterns are more directly reflected in the PPG waveform. In contrast, lower performance is observed for RR, stress, and SDB detection, suggesting that these tasks remain more challenging due to weaker or more indirect physiological signatures within 10-second PPG segments, higher inter-subject variability, and increased sensitivity to acquisition conditions. Overall, these results highlight both the diversity and challenges of \SysName, while motivating future research on more robust multimodal physiological understanding.

\begin{table*}[]
  \centering
  \small
  \caption{In-domain evaluation across different tasks. Here, the PPG encoder used is PulsePPG (except blind).}
  \label{tab:benchmark_main}
  \resizebox{\textwidth}{!}{
  \begin{tabular}{l c c c c c c c c c c c c c}
  \toprule
  \textbf{Model} &
  \textbf{HR} &
  \textbf{BP} &
  \textbf{SDNN} &
  \textbf{RMSSD} &
  \textbf{pNN50} &
  \textbf{RR} &
  \textbf{AF} &
  \textbf{Arrhythmia} &
  \textbf{SpO$_2$} &
  \textbf{SDB} &
  \textbf{Stress} &
  \textbf{SQI} &
  \textbf{Avg. $\pm$ Std} \\
  \midrule

  Blind PPG            & 83.4 & 48.3 & 52.3 & 46.2 & 48.9 & 33.1 & 85.3 & 28.9 & 84.4 & 24.1 & 18.3 & 49.3 & 50.2 $\pm$ 22.3 \\
  Deaf PPG             & 49.1 & 30.6 & 56.9 & 53.0 & 55.0 & 42.4 & 52.3 & 42.6 & 77.9 & 51.0 & 1.7  & 7.8  & 43.4 $\pm$ 20.3 \\
  Fusion PPG           & 74.9 & 59.9 & 73.1 & 54.4 & 69.0 & 39.7 & 84.8 & 65.6 & 72.2 & 43.0 & 41.5 & 85.3 & 63.6 $\pm$ 15.4 \\
  \midrule
  Qwen2.5-7B-Instruct  & 90.0 & 60.0 & 85.3 & 72.3 & 78.2 & 32.7 & 85.4 & 59.6 & 95.4 & 54.1 & 49.7 & 84.7 & \textbf{70.6 $\pm$ 18.4} \\
  \bottomrule
  \end{tabular}
  }
\end{table*}

 \begin{tcolorbox}[
 float,
      colframe=black,
      title=\footnotesize{Instruction-Following Prompt Format used in \textsc{PulseLM}.},
      label={box:pulselm_prompt}
  ]
  \small

  \textbf{[System]}\\
  You are a physiological signal analysis expert specializing in PPG-based clinical classification.\\
  Rules:\\
  \hspace*{4mm}\textbullet\ Answer MUST be exactly one option from the provided list.\\
  \hspace*{4mm}\textbullet\ Output format MUST be strict: \texttt{<answer>OPTION</answer>}\\
  \hspace*{4mm}\textbullet\ Do not output any extra text.

  \vspace{1mm}
  \textbf{[User]}\\
  Task:\\
  \textit{\{Question:\}}

  \vspace{0.5mm}
  Options:\\
  \hspace*{4mm}\textbullet\ \textit{option\textsubscript{1}}\\
  \hspace*{4mm}\textbullet\ \textit{option\textsubscript{2}}\\
  \hspace*{4mm}\textbullet\ \ldots

  \vspace{0.5mm}
  Return ONLY:\\
  \texttt{<answer>OPTION</answer>}

  \vspace{1mm}
  \textbf{[Assistant]}\\
  \texttt{<answer>}\textit{\{\}}\texttt{</answer>}

  \end{tcolorbox}

\section{Discussion}

\subsection{Unifying PPG Inference}
\label{sec:qa_motivation}

In this work, question answering provides a principled framework for linking raw physiological signals to high-level semantic interpretation. Rather than training separate models for each downstream task, a QA formulation allows heterogeneous physiological attributes to be expressed as questions with interpretable answer choices. This mirrors how clinicians and end users naturally reason about physiological data, for example, by asking whether a heart rate is within a normal range, whether a signal is sufficiently clean for interpretation, or whether a recording indicates health risk.

From a modeling perspective, QA provides several advantages: 1) it enables a unified supervision interface across diverse tasks, eliminating the need for task-specific output heads and loss functions; 2) when answer spaces are constrained to categorical choices, QA reduces ambiguity and supports objective evaluation using standard classification metrics; 3) By associating continuous waveform patterns with discrete semantic concepts, QA also encourages models to learn latent physiological representations that are both interpretable and transferable across tasks.

Crucially, a QA-based formulation aligns naturally with multimodal large language models. By conditioning language generation or selection on raw PPG inputs, such models can learn to ground linguistic responses in physiological signals, enabling richer forms of human-machine interaction. However, realizing this potential requires datasets that provide large-scale, physiologically grounded QA supervision across diverse recording conditions.

Motivated by these observations, we design \SysName to reformulate PPG understanding as a question answering problem. By harmonizing heterogeneous public datasets and converting existing physiological annotations into structured QA supervision, \SysName aims to provide the missing foundation for language-enabled PPG modeling. In the following section, we describe the construction of this dataset in detail, including data sources, ground truth harmonization, and QA generation procedures.

\subsection{Limitations and Future Works}
\label{sec:discussion}

Despite its scale and breadth, \SysName has several points that merit discussion. As presented, \SysName is constructed from publicly available datasets, which may exhibit variations in population characteristics and label distributions across tasks. To promote broad coverage, we intentionally incorporate datasets containing a range of physiological conditions, including clinically relevant abnormalities, thereby improving representation beyond standard cases. The released dataset and task are naturally challenging, but further gains would also be achieved through more targeted data curation and stronger integration of domain knowledge into model design. It is worth noting that all source datasets used are de-identified originally, with certain metadata that may still carry external information (e.g., age, gender, BMI, and skin tone). Model performance varies across such factors, and these subgroup effects are not investigated in this dataset work.

A further consideration can be the provenance and clinical fidelity of the QA pairs. Because all questions and answers are derived systematically from structured metadata fields in the source datasets rather than from free-form clinical records or expert textual annotations, the generation process is constrained by the label vocabulary and schema of each dataset. This design choice limits hallucination risk and ensures that every answer is traceable to an original dataset attribute, yet does not always capture the full complexity of real-life clinical practice. \SysName is intended for research purposes on signal-text learning, and models trained on it do not replace clinically validated measurements or expert diagnosis.

At the same time, \SysName opens several promising directions for future work. One natural extension is the incorporation of additional expert-verified annotations and higher-level clinical summaries to further enhance interpretability and linguistic fidelity. Beyond the current QA setting, \SysName also includes structured textual information, which provides a foundation for exploring open-ended physiological report generation. In this setting, models could be trained to produce free-form textual summaries grounded in PPG signals and contextual cues, enabling richer and more flexible interaction than discrete QA alone.

Another important direction is multimodal representation learning, where textual supervision from QA pairs or generated reports is leveraged to guide self-supervised or weakly supervised learning of transferable PPG representations. Such approaches may improve robustness across tasks, datasets, and sensing conditions. Finally, expanding cross-domain generalization benchmarks and developing confidence-aware generation mechanisms remain critical steps toward deploying physiological language models in safety-sensitive health monitoring applications.

\clearpage

\end{document}